\documentclass{bmvc2k}

\usepackage{latexsym}
\usepackage{amsmath}
\usepackage{amssymb}
\usepackage{bm}
\usepackage{url}

\title{%
  Few-shot learning of neural networks from scratch by pseudo example optimization
}

\addauthor{Akisato Kimura}{akisato@ieee.org}{2}
\addauthor{Zoubin Ghahramani}{zoubin@eng.cam.ac.uk}{1}
\addauthor{Koh Takeuchi}{koh.t@acm.org}{2}
\addauthor{Tomoharu Iwata}{iwata.tomoharu@lab.ntt.co.jp}{2}
\addauthor{Naonori Ueda}{ueda.naonori@lab.ntt.co.jp}{2}

\addinstitution{
  Department of Engineering\\
  University of Cambridge\\
  Cambridge, UK.
}
\addinstitution{
  NTT Communication Science Laboratories\\
  Keihanna Science City, Japan.
}
\addinstitution{
  Uber AI Labs\\
  California, USA.
}

\runninghead{Kimura et al.}{Few-shot learning of neural networks from scratch}

\begin{document}

\maketitle

\begin{abstract}
  In this paper, we propose a simple but effective method for training neural networks with a limited amount of training data.
  Our approach inherits the idea of knowledge distillation that transfers knowledge from a deep or wide reference model to a shallow or narrow target model.
  The proposed method employs this idea to mimic predictions of reference estimators that are more robust against overfitting than the network we want to train.
  Different from almost all the previous work for knowledge distillation that requires a large amount of labeled training data, the proposed method requires only a small amount of training data.
  Instead, we introduce pseudo training examples that are optimized as a part of model parameters.
  Experimental results for several benchmark datasets demonstrate that the proposed method outperformed all the other baselines, such as naive training of the target model and standard knowledge distillation.
\end{abstract}

\def\r{\bm{r}}
\def\u{\bm{u}}
\def\x{\bm{x}}
\def\y{\bm{y}}
\def\X{\bm{X}}
\def\Y{\bm{Y}}
\def\bR{\mathbb R}
\newcommand{\todo}[1]{\textcolor{red}{\bf (TODO: {#1})}}

\section{Introduction}
\label{sec:intro}

The current state-of-the-art in computer vision heavily relies on the supervised learning of deep neural networks with large-scale datasets.
However, constructing such large-scale datasets generally requires painstaking effort, and in many real-world applications, only a limited number of training examples can be obtained.
Deep neural networks easily overfit the training data, especially when only a small amount of training data is available.
Several techniques for alleviating overfitting in deep learning have been proposed so far, e.g. semi-supervised learning \cite{Goodfellow2014}, transfer learning \cite{Caruana1994} and few-shot learning \cite{Koch2015}.
However, all the above approaches require either a large number of unsupervised training examples or a pre-trained model trained with a large amount of supervised training data, and thus learning neural networks from only a few examples remains a key challenge.

\begin{figure}[t]
  \begin{center}
    \includegraphics[width=0.75\linewidth]{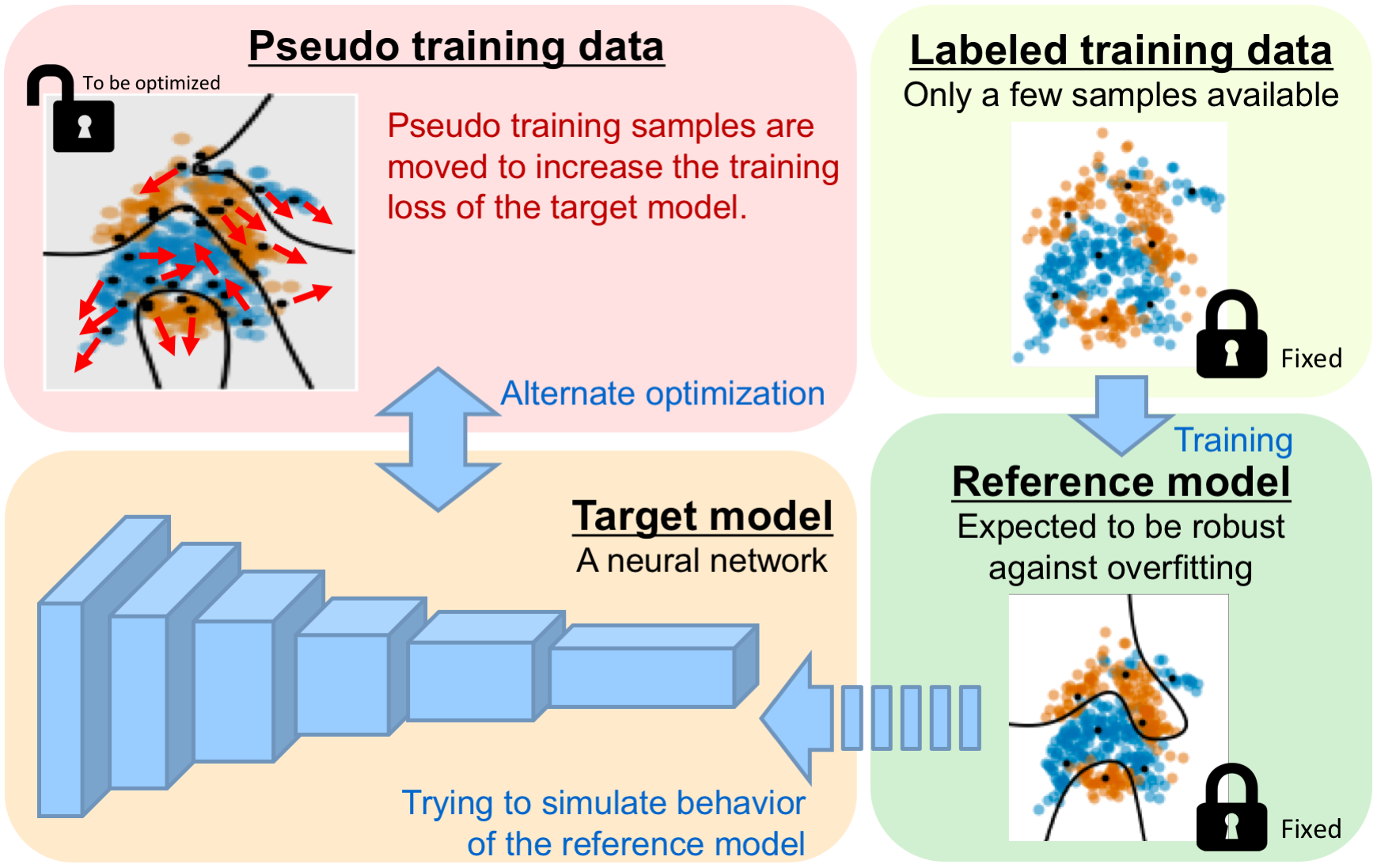}
  \end{center}
  \caption{Basic idea of the proposed method}
  \label{fig:outline}
\end{figure}

Meanwhile, several other estimators such as support vector machines (SVMs) and Gaussian processes (GPs) can ease the adverse effect of overfitting by making use of Bayesian principle or maximum margin.
The universal approximator theorem \cite{Hornik1989} guarantees that an infinitely wide neural network with at least one hidden layer can represent any Lipschitz continuous function to an arbitrary degree of accuracy.
This theorem implies that there exists a neural network that well imitates the behavior of other estimators, while keeping a great representation power of neural networks.

In this paper, we propose a novel method for training neural networks with a small amount of supervised training data.
Figure \ref{fig:outline} shows the basic idea of our proposed method named \emph{imitation networks}.
The proposed method first trains a reference model with a given small amount of supervised training data, and then transfers knowledge from the reference model to a target neural network model in a similar manner to knowledge distillation \cite{Bucilua2006,Ba2014,Hinton2014}.
Although any types of black-box estimators can be applied as a reference model in our method in principle, we particularly select GPs as reference models that can provide local smoothness of predictions.
Different from almost all the previous work for knowledge distillation that employs a large number of supervised training examples, our proposed method requires only a few supervised training examples for knowledge transfer.
To augment training examples, we introduce inducing points \cite{Snelson2005} that are pseudo training examples helping the model training tractable or much easier.
In the original inducing point methods used for scalable GP inference, both inducing points and model parameters are updated to increase the objective function that is actually a lower bound (ELBO) on the marginalized likelihood.
In our proposed method, however, parameters of the target model are updated to \emph{decrease} a training loss, but pseudo training examples are updated to \emph{increase} the training loss.
By doing this, we can move pseudo training examples toward areas where the current target model has not been well trained.
We also introduce fidelity weighting \cite{Dehghani2018} for eliminating harmful pseudo training example based on uncertainty in predictions obtained from reference models.

Our main contributions of this paper can be summarized as follows:
\begin{enumerate}
\item Presenting a novel framework for few-shot training of neural networks based on knowledge transfer.
\item Applying the idea of inducing points into the training of neural networks, which can be optimized in almost the same way as model parameters of neural networks.
\end{enumerate}

\section{Related work}
\label{sec:related}

{\bf Techniques for avoiding overfitting:}
Several techniques have already been proposed for alleviating overfitting in neural network training.
Widely used techniques include data augmentation, model regularization, early stopping, dropout and weight decay, all of which have already been implemented into almost all the deep learning libraries.
Semi-supervised learning is also a major approach when a large amount of unsupervised training examples are available.
Recently, generative models such as generative adversarial networks (GAN) \cite{Goodfellow2014} have become one of the most popular approaches for semi-supervised learning as they can disentangle the supervised information from many other latent factors of variation in a principled way \cite{Kingma2014}.
Transfer learning \cite{Caruana1994} is also widely used for neural network training when a base model trained with a large amount of training data for related tasks is available.
One-shot or few-shot learning \cite{Koch2015,Bertinetto2016,Vinyals2016} is a problem of learning a classifier from only a few supervised examples per class.
As with transfer learning, few-shot learning relies on reference models pre-trained with a large amount of labeled training examples whose class labels are different from target class labels, and captures the characteristics of the target classes from relationships among other reference classes and a single target example.
Compared with those approaches, our method
(1) requires no additional training examples unlike semi-supervised learning, and
(2) achieves few-shot learning \emph{from scratch}, meaning that the reference model is trained with a few training examples, and neither additional examples nor reference models trained with a large amount of supervised training examples are required.

{\bf Knowledge distillation:}
Knowledge distillation is a class of techniques for training a shallow and/or narrow network, and it is also called model distillation or model compression.
More specifically, knowledge distillation learns a small student network by transferring knowledge from large teacher networks, mainly for implementing the network onto devices with limited computational power.
Bucilua et al. \cite{Bucilua2006} pioneered this approach, and Ba and Caruana \cite{Ba2014} extended this idea to deep learning.
Hinton et al. \cite{Hinton2014} generalized the previous methods by introducing a new metric between the output distribution of teacher and student predictions.
Other previous researches demonstrates that knowledge can be transferred from either intermediate layers of a teacher model \cite{Romero2015,Yim2017}, from multiple teacher models \cite{You2017}, mismatched unsupervised stimuli \cite{Kulkarni2017} or examples reconstructed from meta-data \cite{Lopes2017}.
Recent papers \cite{Papernot2015,Furlanello2017,Tarvainen2017} showed that the knowledge distillation framework is applicable to learn a student network from a teacher network with completely the same structure with the student network.
Meanwhile, our method
(1) inherits the idea of knowledge distillation,
(2) exploits an arbitrary black-box estimator as a teacher, in contrast to standard knowledge distillation \footnote{However, this has already been acknowledged in the context of black-box adversarial attacks \cite{Papernot2016}}, and
(3) does not require a large amount of real training data, and instead employs pseudo training data that is optimized during the model training.

{\bf Inducing point methods:}
Inducing point methods \cite{Snelson2005} have been originally developed for scalable GP inference.
Inducing points are pseudo training examples that can help the model training tractable or much easier.
Candela and Rasmussen \cite{Quinonero2005} implies that initial inducing points can be selected from supervised training examples.
Titsias \cite{Titsias2009} suggests a variational approach which provides an objective function for optimizing inducing points.
In this method, both inducing points and covariate function parameters are updated to increase the objective function that is actually the evidence lower bound (ELBO) on the marginalized likelihood.
Hensman et al. \cite{Hensman2013,Hensman2015} introduced stochastic variational inference for improving scalability, where inducing points are regarded as latent variables and marginalized out in the model inference.
Meanwhile, our method
(1) first introduces inducing points into supervised neural network training, and
(2) updates pseudo training examples to \emph{increase} a training loss but updates model parameters to \emph{decrease} the loss, in an adversarial manner.

\section{Model optimization}
\label{sec:frame}

\subsection{Knowledge distillation}
\label{sec:frame:distill}

Before introducing the framework of our proposed method, we first describe knowledge distillation that is the source of our main idea.
Knowledge distillation is a family of methods that transfer the generalization ability of a pre-trained reference model $g(\x; {\bm\theta}_g)$ to a target model $f(\x; {\bm\theta}_f)$, where $\x$ is an input example, and ${\bm\theta}_g$ and ${\bm\theta}_f$ are model parameters of the reference and target models, respectively.
In the following, we will omit model parameters ${\bm\theta}_f$ and ${\bm\theta}_g$ for simplicity unless we explicitly state.
Recent knowledge distillation methods are mainly focused on the training of a shallow or narrow neural network as the target model with the help of reference models that are much deeper or wider neural networks, by minimizing the following distillation loss $L_{\rm dis}$ with respect to the target $f(\cdot)$:
\begin{align}
  L_{\rm dis}(\X^{\rm L}, \Y^{\rm L}) &=
    \frac{\lambda_1}{N_{\rm L}} \sum_{n=1}^{N_{\rm L}} D_1(\y^{\rm L}_n, f(\x^{\rm L}_n))
    + \frac{\lambda_2}{N_{\rm L}} \sum_{n=1}^{N_{\rm L}} D_2(g(\x^{\rm L}_n), f(\x^{\rm L}_n)),
    \label{eq:distillation_loss}
\end{align}
where
$\X^{\rm L}=\{\x^{\rm L}_1, \ldots, \x^{\rm L}_{N_{\rm L}}\}$ is a set of supervised training examples,
$\Y^{\rm L}=\{\y^{\rm L}_1, \ldots, \y^{\rm L}_{N_{\rm L}}\}$ is a set of the corresponding supervisors,
$D_1(\y, \hat{\y})$ is a supervised loss comparing a supervisor $\y$ and a prediction $\hat{\y}$,
$D_2(\hat{\y}_1, \hat{\y}_2)$ is an unsupervised loss computing a loss between two different predictions $\hat{\y}_1$ and $\hat{\y}_2$,
and $\lambda_1$ and $\lambda_2$ are constants balancing two different losses.
Knowledge distillation usually deals with classification problems, where each supervisor $\y^{\rm L}_n$ is a one-hot vector, however, it can be directly applied to regression or other machine learning problems by replacing the supervisors into the shapes appropriate for the problem to be solved.

\subsection{Proposed loss function}
\label{sec:frame:proposed}

As shown in Eq. \eqref{eq:distillation_loss}, almost all the formulations of the previous knowledge distillation work employ supervised training examples $(\X^{\rm L}, \Y^{\rm L})$ to train a target model $f(\cdot)$.
However, you may notice that minimizing the distillation loss with a limited supervised examples causes overfitting, and real supervisions $\Y^{\rm L}$ are not required for computing the unsupervised loss $D_2$.
Based on this observation, our proposed method called \emph{imitation networks} newly introduces pseudo training examples $\X^{\rm P}=(\x^{\rm P}_1,\ldots,\x^{\rm P}_{N_{\rm P}})$ to train a target model $f(\cdot)$ from a reference model $g(\cdot)$, and minimizes the following loss function $L_{\rm imi}$ that we call \emph{the imitation loss}:
\begin{align}
  L_{\rm imi}(\X^{\rm L}, \Y^{\rm L}, \X^{\rm P}) &=
    \frac{\lambda_1}{N_{\rm L}} \sum_{n=1}^{N_{\rm L}} D_1(\y^{\rm L}_n, f(\x^{\rm L}_n))
    + \frac{\lambda_2}{N_{\rm P}} \sum_{n=1}^{N_{\rm P}} D_2(g(\x^{\rm P}_n), f(\x^{\rm P}_n)).
    \label{eq:imitation_loss}
\end{align}

In the same way as knowledge distillation, a reference model $g(\cdot)$ is first pre-trained with supervised training examples $(\X^{\rm L}, \Y^{\rm L})$.
However, we have to note that only a few number of supervised examples are available in our problem setting.
A reference model $g(\cdot)$ can be a single estimator or an ensemble of multiple estimators.
We can build multiple estimators from a single model by changing hyper-parameters, such as a variance and a length scale in RBF kernels for GPs.
When introducing an ensemble of multiple estimators, we choose to average their predictions in a similar manner to the previous work \cite{Hinton2014}.

The above formulation indicates that the target model tries to imitate predictions of the reference model for pseudo examples and at the same time it tries to return predictions for supervised examples as accurately as possible.
Therefore, the resulting target model is not a copy of the reference model and can inherit two different properties coming from the reference model and neural networks.
In particular for the use of Gaussian processes as referencd models, they introduce local smoothness of predictions into neural networks, which is lacked in standard neural networks \cite{Bradshaw2017} and is proved to be effective for overfitting \cite{Miyato2016}.

\subsection{Fidelity weighting}
\label{sec:frame:weight}

We note that all the pseudo training examples are not always useful for knowledge transfer.
Examples yielding unreliable predictions from the reference model are more likely harmful and should be discarded in the model training.
For this purpose, we introduce fidelity weighting \cite{Dehghani2018}, which adaptively weighs training examples based on uncertainty of predictions obtained from the reference model.
With the introduction of fidelity weighting, the imitation loss Eq. \eqref{eq:imitation_loss} can be replaced by the following equation.
\begin{align}
  L_{\rm imi}(\X^{\rm L}, \Y^{\rm L}, \X^{\rm P}) &=
    \frac{\lambda_1}{N_{\rm L}} \sum_{n=1}^{N_{\rm L}} D_1(\y^{\rm L}_n, f(\x^{\rm L}_n))
    + \frac{1}{N_{\rm P}} \sum_{n=1}^{N_{\rm P}} \lambda_2(g, \x^{\rm P}_n) D_2(g(\x^{\rm P}_n), f(\x^{\rm P}_n)).
    \label{eq:imitation_loss_fidelity}
\end{align}
where $\lambda_2(g, \x^{\rm P}_n)$ is a new example-wise weight for a pseudo example $\x^{\rm P}_n$, which is computed from uncertainty $\sigma_g(\x^{\rm P}_n)$ of the reference prediction $g(\x^{\rm P}_n)$, as follows:
\begin{align}
  \lambda_2(g, \x^{\rm P}_n) &=
    \hat{\lambda}_2 \exp(- \log(\hat{\lambda}_2 / \overline{\lambda}_2) \sigma_g(\x^{\rm P}_n) / \overline{\sigma}_g),
    \label{eq:fidelity_weighting}
\end{align}
where $\hat{\lambda}_2$ is an upper bound of weights, $\overline{\sigma}_g$ is a mean uncertainty over all the pseudo examples and $\overline{\lambda}_2$ is a weight for the mean uncertainty.
Fidelity weighting is valid for various kinds of reference estimators, unless Bayesian treatments cannot be applied to the reference estimators in principle.
For example, the original work \cite{Dehghani2018} utilized GP classifiers.
Bayesian variants of SVMs \cite{Zhang2006} and Bayesian neural networks \cite{Gal2016} can be employed for this purpose.

\section{Pseudo example optimization}
\label{sec:pseudo}

When transferring knowledge useful for improving performance, the selection of pseudo examples plays a significant role.
We describe this step in the this section.

\subsection{Inducing point method}
\label{sec:pseudo:induce}

As described in Section \ref{sec:related}, the inducing point method has been originally developed for scalable GP inference, and in this method both inducing points and model parameters are updated to increase the objective function.
On the other hand, in our method, updates of model parameters and pseudo training examples are going off in completely different directions.
Namely, model parameters are updated to \emph{decrease} the imitation loss, meanwhile pseudo examples are updated to \emph{increase} the loss.
By doing this, we can move pseudo examples towards areas where the current target model has not been well trained.

Our proposed technique for updating pseudo examples is inspired by adversarial training \cite{Szegedy2015}, which increases the robustness of neural networks to adversarial examples, formed by applying small but intentionally worst-case perturbations to examples from the dataset.
The most popular technique for generating an adversarial example $\x_{\rm AT}(\x^{\rm L}, \y^{\rm L})$ from a given supervised example $(\x^{\rm L}, \y^{\rm L})$ is a fast gradient sign method \cite{Goodfellow2015}.
\begin{align}
  \x_{\rm AT}(\x^{\rm L}, \y^{\rm L}) &= \x^{\rm L} + \epsilon~ {\rm sign}\{\nabla_{\x^{\rm L}} D_1(\y^{\rm L}, f(\x^{\rm L}))\},
    \label{eq:adversarial_example}
\end{align}
where $\nabla_{\x}$ is a partial derivative with respect to $\x$ and $\epsilon\ge 0$ is a small constant.

You may notice that the adversarial example $\x_{\rm AT}(\x, \y)$ can be seen as a stochastic update of the original example $\x$ to \emph{increase} the loss and a small constant $\epsilon$ can be seen as a learning rate, if the gradient sign part ${\rm sign}\{\nabla_{\x} D_1(\y, f(\x))\}$ is replaced by a standard stochastic gradient $\nabla_{\x} D_1(\y, f(\x))$.
Also, the ground-truth supervisor $\y$ can be removed from the update, since every pseudo training example $\x^{\rm P}$ has a soft supervision $g(\x^{\rm P})$ instead.
Based on the above discussion, the update of a pseudo training example $\x^{\rm P}$ can be obtained as
\begin{align}
  \x_{\rm imi}(\x^{\rm P}) &= \x^{\rm P} + \epsilon\nabla_{\x^{\rm P}} D_2(g(\x^{\rm P}), f(\x^{\rm P})).
    \label{eq:pseudo_update}
\end{align}
Instead of the direct use of this stochastic gradient update, we can introduce the recent advances of stochastic optimization such as Adam \cite{Kingma2015} and Nadam \cite{Dozat2016} for faster optimization.
In addition, the stochastic gradient part can be replaced by other types of adversarial examples, such as natural \cite{Zhao2018} and spatially transformed adversary \cite{Xiao2018}.

We can build an adversarial training procedure that alternatively updates parameters ${\bm\theta}_f$ of the target model to \emph{decrease} the imitation loss and pseudo examples $\X^{\rm P}$ to \emph{increase} the loss.
However, such an adversarial training procedure is well known to be unstable \cite{Salismans2016}.
Instead, we adopt another approach that augments pseudo examples.
We employ a fixed set $\X^{{\rm P}(t)}$ of pseudo examples for the $t$-th step of model training, and update another set $\X^{{\rm P}(t+1)}$ of pseudo examples that is originally a carbon copy of the current set $\X^{{\rm P}(t)}$.
After the convergence of the $t$-th training step, the current set $\X^{{\rm P}(t)}$ is merged into the next set $\X^{{\rm P}(t+1)}$.
Holding pseudo examples on fixed positions will make the model training more stable, and augmenting pseudo examples whose predictions are distant from those of the reference model makes the target model to be closer to the reference model.

\subsection{Algorithm}
\label{sec:pseudo:algo}

Summarizing the above discussion, the training procedure of our proposed imitation networks can be described as follows:
\begin{enumerate}
\item A reference model $g(\cdot)$ is trained with (a few) labeled examples $(\X^{\rm L}, \Y^{\rm L})$.
\item An initial set $\X^{{\rm P}(0)}$ of pseudo examples is generated, a carbon copy $\X^{{\rm P}(1)}$ of the initial set $\X^{{\rm P}(0)}$ is created and the index $t$ representing the current training step is set to $0$.
\item Model parameters ${\bm\theta}_f$ of a target model $f(\cdot)$ are updated by using the current pseudo examples $\X^{{\rm P}(t)}$ and their outputs of the reference model $g(\cdot)$ so as to decrease the imitation loss Eq. \eqref{eq:imitation_loss_fidelity}.
\item Every pseudo example in the next set $\X^{{\rm P}(t+1)}$ is updated with Eq. \eqref{eq:pseudo_update} and the current target model $f(\cdot)$.
\item Repeat 3.-4. until a pre-defined number of training epochs are completed.
\item A carbon copy $\X^{{\rm P}(t+2)}$ of the next set $\X^{{\rm P}(t+1)}$ is newly created for further updates
  (i.e. $\X^{{\rm P}(t+2)} \leftarrow \X^{{\rm P}(t+1)}$),
  the current set $\X^{{\rm P}(t)}$ is included into the next set $\X^{{\rm P}(t+1)}$
  (i.e. $\X^{{\rm P}(t+1)} \leftarrow \X^{{\rm P}(t)} \cup \X^{{\rm P}(t+1)}$),
  and the time index $t$ is incremented as $t\leftarrow t+1$.
\item Repeat 3.-6. until a pre-defined number of training steps are completed.
\end{enumerate}

\section{Experiments}
\label{sec:exp}

\subsection{Qualitative analysis}
\label{sec:exp:banana}

\begin{figure*}[t]
  \begin{center}
    \includegraphics[width=0.985\linewidth]{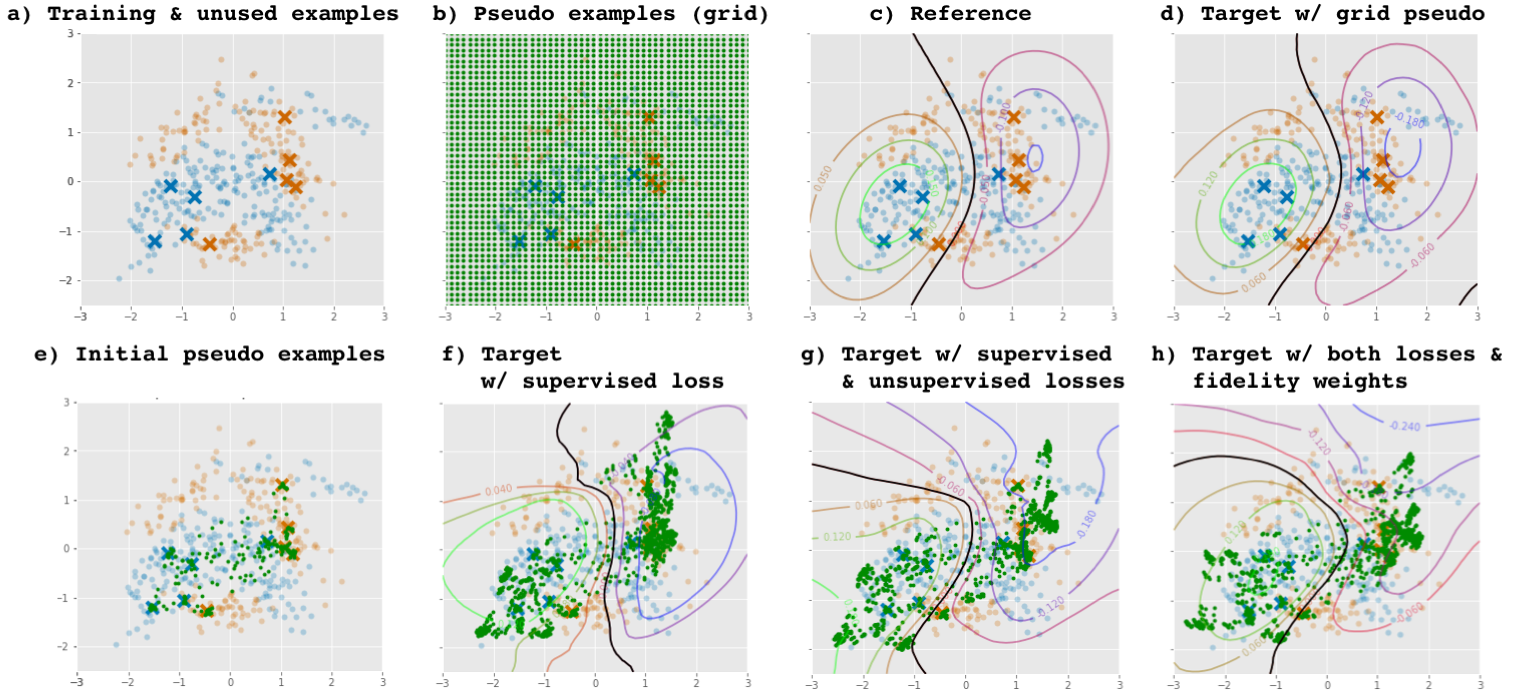}
  \end{center}
  \caption{%
    Visualizations of predictions, where
    orange \& blue crosses = supervised examples,
    orange \& blue dots = other unused examples in the dataset,
    green dots = pseudo examples,
    and black lines = classification boundaries.
    From the top left to the bottom right,
    a) a distribution of examples,
    b) densely distributed pseudo examples,
    c) a reference model trained with supervised examples,
    d) a target model trained with densely distributed pseudo examples,
    e) pseudo examples generated from supervised examples as seeds,
    f) a target model trained with generated pseudo examples,
    g) a target model trained with supervised and pseudo examples.
    h) a target model trained with supervised and pseudo examples + fidelity weighting.
  }
  \label{fig:exp:banana}
\end{figure*}

To check the behavior of the proposed method qualitatively, we first turned to the Banana dataset \footnote{\url{https://github.com/GPflow/GPflow/tree/master/doc/source/notebooks/data}} \cite{Hensman2015} for binary classification with 400 two-dimensional examples.
We embedded each of two-dimensional examples into a 100-dimensional vector space by a linear transformation, where a transformation matrix was randomly determined in advance, and we selected five supervised training examples per class randomly.
The selected supervised examples and other unused examples are shown in Figure \ref{fig:exp:banana} a), where crosses are supervised examples, dots are unused examples in the dataset and a color corresponds to its class label.
We employed a GP classifier with an RBF kernel as a reference model, and built an ensemble of multiple reference models trained with different initial kernel parameters.
We used GPflow \cite{GPflow} with variational Gaussian approximation \cite{Opper2009} and L-BFGS-B optimization \cite{Zhu1997} for GP inference, and utilized mean predictions as pseudo supervisions.
The target model we selected was a $7$-layer fully-connected neural network where each of the intermediate layers has $1000$ units.
We used Nadam \cite{Dozat2016} for model parameter optimization and Adam \cite{Kingma2015} for pseudo example optimization, and their initial learning rates were set to $0.001$ and $0.05$, respectively.
The batch size was $100$, and the number of training epochs was $200$.

First, we examined an extreme case where $2500$ pseudo examples are densely distributed onto whole the feature space.
Although this is not realistic especially for high-dimensional spaces, it is useful to check whether or not our proposed method can imitate the behavior of the reference model precisely.
Figure \ref{fig:exp:banana} b) shows its distribution, where green dots are pseudo examples.
In this case, we did not optimize pseudo examples, and we did not employ fidelity weighting shown in \ref{sec:frame:weight}, which means that we utilized the original imitation loss shown in Eq. \eqref{eq:imitation_loss}.
We utilized Kullback-Leibler divergence for the soft loss $D_2$ and ignored the hard loss $D_1$ (i.e. $\lambda_1=0$).
Figure \ref{fig:exp:banana} c) and d) show decision boundaries of the reference and target models, respectively.
This result indicates that our method imitated the predictions of the reference model almost completely, as expected.

Next, we considered a more realistic setup, where $250$ initial pseudo examples are generated by a mixture of interpolation and Gaussian augmentation of supervised examples, and augmented them to $1000$ with our technique shown in the last part of Section \ref{sec:pseudo:induce}, where the number of pseudo examples is much smaller than the first setup.
Thus, we had $4$ training steps each of which had $50$ training epochs and employed $250$, $500$, $750$ and $1000$ pseudo examples, respectively.
We utilized hinge loss for the hard loss $D_1$ and Kullback-Leibler divergence for the soft losses $D_2$, and both of the weights $\lambda_1$ and $\overline{\lambda}_2$ were set to $1.0$.
Figure \ref{fig:exp:banana} e) shows an initial distribution of pseudo training examples, and f), g) and h) show decision boundaries and optimized pseudo examples of the target models trained with only soft supervisions, with soft and hard supervisions, and supervisions plus fidelity weighting, respectively.
The results indicate that the target model well imitated the reference model even with sparsely distributed pseudo examples.
Also, Figure \ref{fig:exp:banana} g) and h) indicate that the introduction of hard supervisions yielded a better classification performance for supervised examples (see the orange cross at the center bottom) while preserving the decision boundary as much as possible.
A distribution of optimized pseudo examples partly describes properties of our proposed method.
Optimized pseudo examples were distributed over whole the feature space so as to decrease the difference between the target and reference models.

\subsection{Quantitative evaluations}
\label{sec:exp:classify}

\begin{table}[t]
  \footnotesize
  \caption{%
    Classification performances for MNIST (top) and fashion MNIST (bottom) datasets,
    where ``imitation'', ``optimize'' and ``fidelity'' stand for the use of the imitation loss, pseudo example optimization,
    and fidelity weighting, respectively
  }
  \label{table:exp:imitation}
  \begin{center}
    \begin{tabular}{|l||r|r|r|r|r|}\hline
      \#labeled & 10                & 20   & 50   & 100  & 200  \\ \hline\hline
      NN                            & 37.9 & 46.0 & 66.0 & 78.3 & {\bf 86.7} \\ \hline
      GP                            & 39.9 & 51.6 & 64.6 & 73.2 & 80.0 \\ \hline
      Imitation                     & 43.5 & 51.2 & 67.7 & 78.1 & 86.1 \\ \hline
      Imitation, optimize           & 44.1 & 53.7 & 70.0 & 79.5 & {\bf 86.7} \\ \hline
      Imitation, optimize, fidelity & {\bf 44.1} & {\bf 53.9} & {\bf 70.4} & {\bf 80.0} & 86.6 \\ \hline
    \end{tabular}
    \\\vspace{2mm}
    \begin{tabular}{|l||r|r|r|r|r|}\hline
      \#labeled & 10                & 20   & 50   & 100  & 200  \\ \hline\hline
      NN                            & 39.3 & 47.9 & 58.3 & 64.9 & 71.3 \\ \hline
      GP                            & 44.6 & 52.4 & 59.9 & 65.7 & 71.4 \\ \hline
      Imitation                     & 43.6 & 50.9 & 60.0 & 67.3 & {\bf 72.5} \\ \hline
      Imitation, optimize           & 41.2 & 49.7 & 60.1 & 67.3 & 72.2 \\ \hline
      Imitation, optimize, fidelity & {\bf 44.8} & {\bf 52.7} & {\bf 62.1} & {\bf 68.0} & {\bf 72.5} \\ \hline
    \end{tabular}
  \end{center}
\end{table}

Next, we quantitatively evaluated classification performances of the proposed method for several benchmark datasets.
We used MNIST \cite{mnist} and fashion MNIST \cite{Xiao2017} as the datasets for this experiment.
We again employed a GP classifier as a reference model, where all the setups related to the reference models were the same as Section \ref{sec:exp:banana}.
The target model was a 3-layer CNN for both of the datasets, where the detailed configulation can be found in the supplementary material.
We again used Nadam for model parameter optimization and Adam for pseudo example optimization, and both of the initial learning rates were set to $0.02$.
We prepared $1.25K$ initial pseudo examples by interpolating two different supervised examples, and augmented them to $10K$ by employing our technique shown in Section \ref{sec:pseudo:induce}.
Thus, whole the training process contained $8$ training steps and each training step had $25$ training epochs.
The weight $\lambda_1$ for the hard loss was set to $1.0$, and the mean weight $\overline{\lambda}_2$ for the soft loss was decreased $100$ to $1$ as the number of supervised examples increased.
We trained the target model with $1$ to $20$ supervised examples per class randomly selected from the $50$K training examples in each of the datasets.
Then, we tested the trained model with the $10$K test examples, and compared classification accuracy averaged over $20$ different selection of supervised examples.
All the other experimental conditions were the same as the second case in Section \ref{sec:exp:banana}.
We compared the performance of the following $5$ training strategies, namely
(1) the reference estimators,
(2) a naive training of the target neural network,
(3) our method with only the imitation loss (no pseudo example optimization and fidelity weighting),
(4) our method with the imitation loss and pseudo example optimization (no fidelity weighting), and
(5) our method with the imitation loss, pseudo example optimization and fidelity weighting.

Table \ref{table:exp:imitation} shows the experimental result.
The result indicates that our proposed method outperformed a naive training of the target neural network model.
The result also indicates that a naive training of the target neural network outperformed the reference models when a large number of supervised examples of MNIST dataset were used.
Even with this undesirable setting, our proposed method transferring knowledge of a rather weak reference model to the large target model was superior or comparable to the naive training.
In addition, our proposed method outperformed the reference models.
This is because our imitation loss enables us to inherit the properties of two different models, namely a Gaussian process classifier and a neural network, and a Gaussian process classifier as a reference model introduces local smoothness of predictions into neural networks.
Our proposed method reasonably worked well even without pseudo example optimization, however, the introduction of pseudo example optimization improved the classification performance, especially for MNIST dataset.
Although fidelity weighting also improved the performance in some cases, the contribution was minor compared with pseudo example optimization that is our main contribution of this paper.

\section{Conclusion}
\label{sec:conc}

In this paper, we have proposed a simple but effective method for training neural networks with a limited number of training examples.
Our proposed method employed GP as a reference, and built a target neural network so as to imitate the behavior of the reference trained with the limited examples.
%
%
We introduced pseudo examples and optimized them through the process of target model training.
%
%
Since our proposed framework shown in Section \ref{sec:frame:proposed} is generic, it can be directly applied to other combinations of reference and target models.
For example, it is possible to train a deep target network with a pre-trained shallow reference network trained on a small amount of data, which is the inverse of knowledge distillation, which may provide a new way of pre-training deep neural networks.
Meanwhile, our method for optimizing pseudo examples is rather specific, and sophisticated management of pseudo examples remains largely to be investigated.
%
%
%
%


\bibliographystyle{bmvc2k}
\bibliography{bmvc18_imitation}

\end{document}